\documentclass[runningheads]{llncs}
\usepackage{graphicx}
\usepackage{amsmath}
\usepackage{subfigure}
\usepackage{booktabs}
\usepackage{amsfonts,amssymb}
\usepackage[misc]{ifsym}
\begin{document}

\title{Unsupervised Learning of Local Discriminative Representation for Medical Images}
\titlerunning{Local Discriminative Representation for Medical Images}

%

\author{Huai Chen\inst{1}\orcidID{0000-0002-1815-1486}
	\and Jieyu Li\inst{1}
	\and Renzhen Wang\inst{2}
	\and Yijie Huang\inst{1}
	\and Fanrui Meng\inst{1}
	\and Deyu Meng\inst{2}
	\and Qing Peng\inst{3}
	\and Lisheng Wang\inst{1}\orcidID{0000-0003-3234-7511}$^*$$^{(\textrm{\Letter})}$
}
\authorrunning{H. Chen et al.}

%

\institute{Institute of Image Processing and Pattern Recognition, Department of Automation, Shanghai Jiao Tong University, Shanghai, 200240, P. R. China. \email{lswang@sjtu.edu.cn}
\and Xi'an Jiaotong University, Xi'an, P. R. China.
\and Department of Ophthalmology, Shanghai Tenth People’s Hospital, Tongji University, Shanghai,P. R. China. }

\maketitle              
\begin{abstract}
Local discriminative representation is needed in many medical image analysis tasks such as identifying sub-types of lesion or segmenting detailed components of anatomical structures. However, the commonly applied supervised representation learning methods require a large amount of annotated data, and unsupervised discriminative representation learning distinguishes different images by learning a global feature, both of which are not suitable for localized medical image analysis tasks. In order to avoid the limitations of these two methods, we introduce local discrimination into unsupervised representation learning in this work. The model contains two branches: one is an embedding branch which learns an embedding function to disperse dissimilar pixels over a low-dimensional hypersphere; and the other is a clustering branch which learns a clustering function to classify similar pixels into the same cluster. These two branches are trained simultaneously in a mutually beneficial pattern, and the learnt local discriminative representations are able to well measure the similarity of local image regions. These representations can be transferred to enhance various downstream tasks. Meanwhile, they can also be applied to cluster anatomical structures from unlabeled medical images under the guidance of topological priors from simulation or other structures with similar topological characteristics. The effectiveness and usefulness of the proposed method are demonstrated by enhancing various downstream tasks and clustering anatomical structures in retinal images and chest X-ray images.

\end{abstract}

\section{Introduction}
In medical image analysis, transferring pre-trained encoders as initial models is an effective practice, and supervised representation learning is widely applied, while it usually depends on a large amount of annotated data and the learnt features might be less efficient for new tasks differing from original training task \cite{dosovitskiy2015discriminative}. Thus, some researchers turn to study unsupervised representation learning \cite{mahmood2020whole,xie2020instance}, and particularly unsupervised discriminative representation learning was proposed to measure similarity of different images \cite{he2020momentum,wu2018unsupervised,ye2019unsupervised}. However, these methods mainly learn the instance-wise discrimination based on global semantics, and cannot characterize the similarities of local regions in image. Hence, they are less efficient for many medical image analysis tasks, such as lesion detection, structure segmentation, identifying distinctions between different structures, in which local discriminative features are needed to be captured. In order to make unsupervised representation learning suitable for these tasks, we introduce local discrimination into unsupervised representation learning in this work.

It is known that medical images of humans contain similar anatomical structures, and thus pixels can be classifying into several clusters based on their context. Based on such observations, a local discriminative embedding space can be learnt, in which pixels with similar context will distribute closely and dissimilar pixels can be dispersed. In this work, a model containing two branches is constructed following a backbone network, in which an embedding branch is used to generate pixel-wise embedding features and a clustering branch is used to generate pseudo segmentations. Through jointly updating these two branches, pixels belonging to the same cluster will have similar embedding features and different clusters will have dissimilar ones. In this way, local discriminative features can be learnt in an unsupervised way, which can be used for evaluating similarity of local image regions.

The proposed method is further applied to several typical medical image analysis tasks respectively in fundus images and chest X-ray images: (1) The learnt features are utilized in 9 different downstream tasks via transfer learning, including segmentations of retinal vessel, optic disk (OD) and lungs, detection of haemorrhages and hard exudates, etc., to enhance the performances of these tasks. (2) Inspired by specialists' ability of recognizing anatomical structures based on prior knowledge, we utilize the learnt features to cluster local regions of the same anatomical structure under the guidance of topological priors, which are generated by simulation or from other structures with similar topology.

\section{Related work}
Instance discrimination learning method \cite{wu2018unsupervised,he2020momentum,bachman2019learning,ye2019unsupervised} is an unsupervised representation learning framework providing a good initialization for downstream tasks and it can be considered as an extension of exemplar convolution neural network (CNN) \cite{dosovitskiy2015discriminative}. The main conception of instance discrimination is to build an encoder to dispersedly embed training samples over a hypersphere \cite{wu2018unsupervised}. Specifically speaking, a CNN is trained to project each image onto a low-dimensional unit hypersphere, in which the similarity between images can be evaluated by cosine similarity. In this embedding space, dissimilar images are forced to be separately distributed and similar images are forced to be closely distributed. Thus, the encoder can make instance-level discrimination. Wu et al. \cite{wu2018unsupervised} introduce a memory bank to store historical feature vectors for each image. Then the probability of image being recognized as $i$-th example can be expressed by inner product of the embedding vector and vectors stored in the memory bank. And the discrimination ability of encoder is obtained by learning to correctly classify image instance into the corresponding record in the memory bank. However, the vectors stored in the memory bank are usually outdated caused by discontinuous updating. To address this problem, Ye et al. \cite{ye2019unsupervised} propose a framework with siamese network which introduces augmentation invariant into the embedding space to cluster similar images to realize real-time comparison.

The ingenious design enables instance discrimination effectively utilize unlabeled images to train a generalized feature extractor for downstream tasks and shrink the gap between unsupervised and supervised representation learning \cite{he2020momentum}. However, summarizing a global feature for image instance miss local details, which are crucial for medical image tasks, and the high similarity of global semantics between images of same body part makes instance-wise discrimination less practical. Therefore, it is more convinced to focus on local discrimination of medical images. Meanwhile, medical images of the same body part can be divided into several clusters due to the similar anatomical structures, which inspires us to propose a framework to cluster similar pixels to learn local discrimination.

\begin{figure*}[b]
	\centering	
	\includegraphics[width=\textwidth]{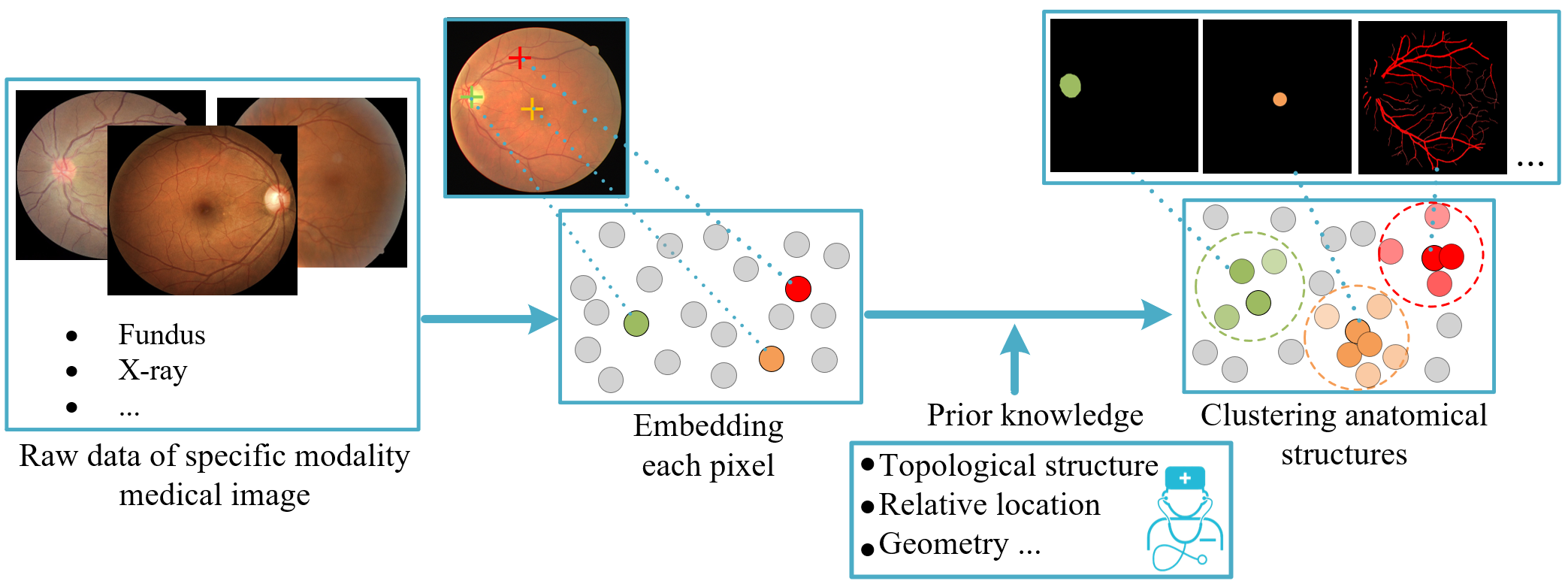}
	\caption{Illustration of our proposed learning model.} 
	\label{fig:framework}
\end{figure*}
\section{Methods}

The illustration of our unsupervised framework is shown in Figure \ref{fig:framework}. This model has two main components. The first is learning a local discriminative representation, which aims to project pixels into a $l_2$-normalized low-dimensional space, i.e. a $K$-D unit hypersphere, and pixels with similar context should be closely distributed and dissimilar pixels should be far away from each other on this embedding space. The learnt local discriminative representation can be taken as a good feature extractor for downstream tasks. The second is introducing prior knowledge of topological structure and relative location into local discrimination, where the prior knowledge will be fused into the model to make the distribution of pseudo segmentations closer to the distribution of priors. By combining priors of structures with local discrimination, regions of the expected anatomical structure can be clustered.

\subsection{Local discrimination learning}
\begin{figure*}[t]
	\centering	
	\includegraphics[width=\textwidth]{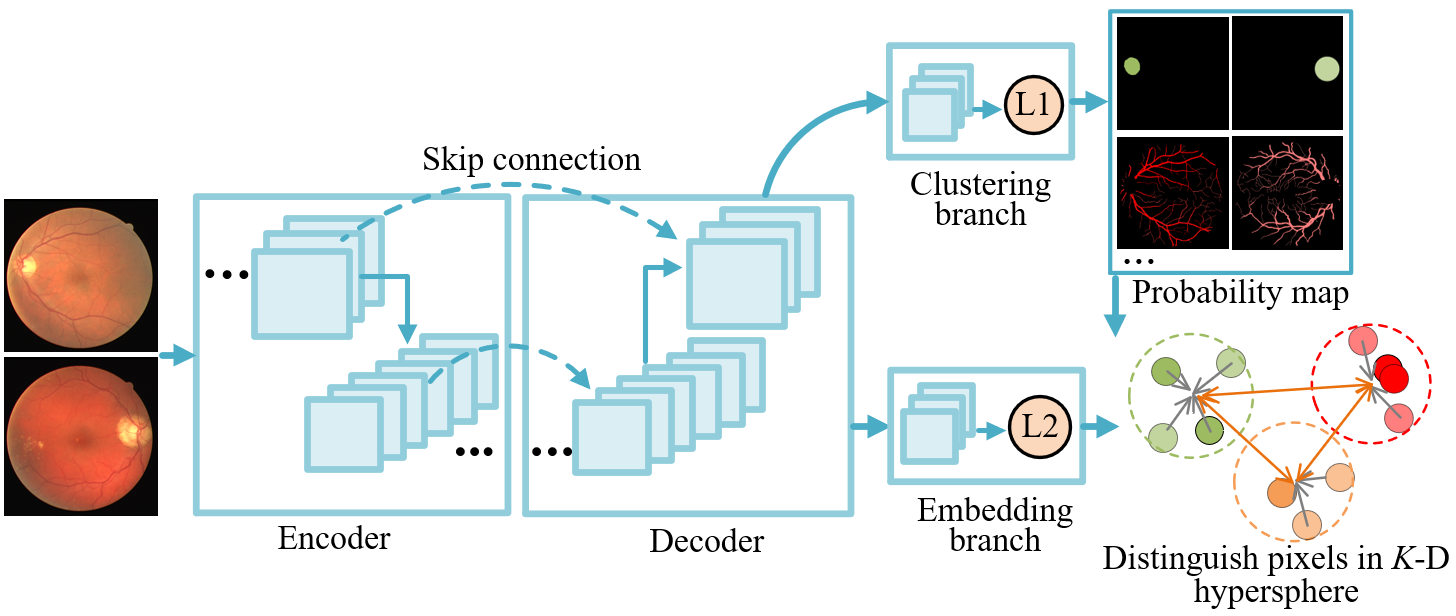}
	\caption{Illustration of local discrimination learning.} 
	\label{fig:learning_region_discrimination}
\end{figure*}
As medical images of the same body region contain same anatomical structures, image pixels can be classified into several clusters, each of which corresponds to a specific kind of structure. Therefore, local discrimination learning is proposed to train representations to embed each pixel onto a hypersphere, on which pixels with similar context will be encoded closely. To achieve this, two branches, including an embedding branch to encode each pixel and a clustering branch to generate pseudo segmentations to cluster pixels, are built following a backbone network and trained in a mutually beneficial manner.

\subsubsection{Notation:} We denote $f_\theta$ as the deep neural network, where $\theta$ is the parameters of network. The unlabeled image examples are denoted as $X=\{x_{1},...,x_{N}\}$ where $x_{i}\in \mathbb{R}^{H,W}$. After feeding $x_{i}$ into the network, we can get embedding features $v_{i}$ and probability map $r_{i}$, i.e., $v_{i},r_{i}=f_\theta (x_{i})$, where $v_{i}(h,w) \in \mathbb{R}^{K}$ is the $K$-dimensional encoded vector for position $(h,w)$ of image $x_{i}$ and $r_{i}(h,w) \in \mathbb{R}^{M}$ is a vector representing the probability of classifying pixel $x_{i}(h,w)$ into $M$ clusters. And $r_{mi}(h,w)$ denotes the probability of classifying pixel $x_{i}(h,w)$ into the $m$-th cluster. We force $||r_{i}(h,w)||_{1}=1$ and $||v_{i}(h,w)||_{2}=1$ by respectively setting $l_{1}$ and $l_{2}$ normalization in clustering branch and embedding branch.

\subsubsection{Jointly train clustering branch and embedding branch:} 
After getting embedding features and pseudo segmentations, the center embedding feature $c_{m}$ of $m$-th cluster can be formulated as followed:
\begin{equation}
	c_{m}=\frac{\sum_{i,h,w}{r_{mi}(h,w)v_{i}(h,w)}}{||\sum_{i,h,w}{r_{mi}(h,w)v_{i}(h,w)}||_{2}}
\end{equation}
Where $l_{2}$ normalization is used to make $c_{m}$ on the hypersphere. Thus, the similarity between $c_{m}$ and $v_{i}(w,h)$ can be evaluated by cosine similarity as followed:
\begin{equation}
	t(c_{m},v_{i}(w,h))=c_{m}^{T}v_{i}(w,h)
\end{equation} 
To make pixels of same cluster closely distributed and pixels of different clusters dispersedly distributed, there should be high similarity between $v_i(w,h)$ and corresponding center embedding features $c_{m}$, and low similarity between $c_{m}$ and $c_{n}(m\neq n)$ as well. Thus, the loss function can be formulated as followed:
\begin{equation}
	loss_{ld}=-\frac{1}{MNHW}\sum_{m,i,h,w}r_{mi}(h,w)t(c_{m},v_{i}(w,h))+\frac{1}{M(M-1)}\sum_{m,n \neq m}c_{m}^{T}c_{n}
\end{equation}

\subsubsection{More constraints:} 
We also add entropy loss and area loss to make high confidence of predictions and avoid blank outputs for some clusters. The losses are as followed:
\begin{equation}
	loss_{entropy}=-\frac{1}{MNHW}\sum_{m,i,h,w}r_{mi}(h,w)logr_{mi}(h,w)
\end{equation}
\begin{equation}
	area_{mi}=\sum_{h,w}r_{mi}(h,w)
\end{equation}
\begin{equation}
	loss_{area}=\frac{1}{NM}\sum_{m,i}relu(\frac{1}{4M}HW-area_{mi})
\end{equation}
Where $relu$ is rectified linear units \cite{glorot2011deep}, $loss_{area}$ will impose punishment if the area of pseudo segmentation is smaller than $\frac{1}{4M}HW$.

\subsection{Prior-guided anatomical structure clustering}
\begin{figure*}[t]
	\centering	
	\includegraphics[width=\textwidth]{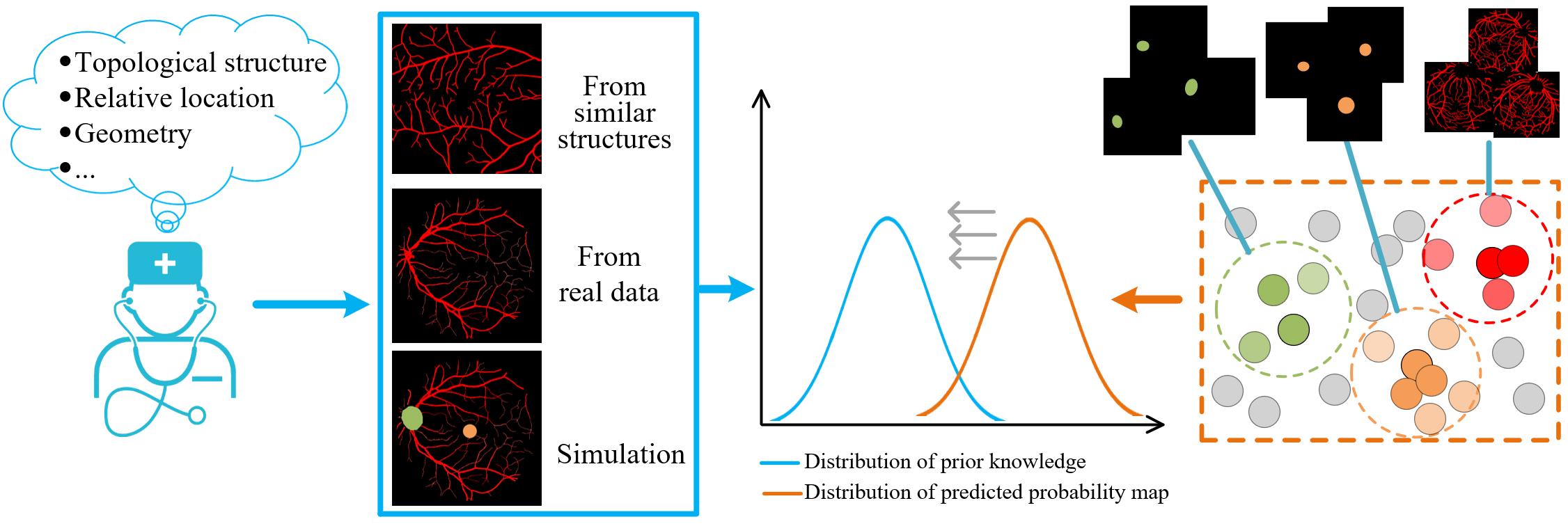}
	\caption{Based on prior knowledge and local discrimination to recognize structures.} 
	\label{fig:prior_knowledge}
\end{figure*}
Commonly, specialists can easily identify anatomical structures based on corresponding prior knowledge, including relative location, topological structure, and even based on knowledge of similar structures. Therefore, DNN's ability of recognizing structures based on local discrimination and topological priors is studied in this part. Reference images, which are binary masks of similar structures, real data or simulation and show knowledge of location and topological structure, is introduced to the network to force the clustering branch to obtain corresponding structures as shown in Figure \ref{fig:prior_knowledge}.

We denote the distribution of $m$-th cluster as $P_{m}$ and the distribution of corresponding references as $Q_{m}$. The goal of optimization is to minimize Kullback-Leibler (KL) divergence between them, and it can be formulated as followed:
\begin{equation}
	\mathop{min}_{f_{\theta}}KL(P_{m}||Q_{m})=\sum_{i}P_{m}(r_{mi})log\frac{P_{m}(r_{mi})}{Q_{m}(r_{mi})}
\end{equation}
To minimize the KL divergence between $P_{m}$ and $Q_{m}$, adversarial learning \cite{goodfellow2014generative} is utilized to encourage the produced pseudo segmentation to be similar as the reference mask. During training, a discriminator $D$ is set to discriminate pseudo segmentation $r_{m}$ and reference mask $s_{m}$, while $f_{\theta}$ aims to cheat $D$. The loss function for $D$ and adversarial loss for $f_{\theta}$ are defined as followed:
\begin{equation}
	loss_{D}=loss_{bce}(D(s_{m}),1)+loss_{bce}(D(r_m),0)
\end{equation}
\begin{equation}
	loss_{bce}(\hat{y},y)=-\frac{1}{N}\sum_{i}(y_ilog\hat{y}_{i}+(1-y_{i})log(1-\hat{y}_i))
\end{equation}
\begin{equation}
	loss_{adv}=loss_{bce}(D(r_m),1)
\end{equation}

\subsubsection{Reference masks:} (1) From similar structures: Similar structures share similar geometry and topology. Therefore, we can utilize segmentation annotations from similar structures to guide the segmentation of target, e.g., annotations of vessel in OCTA can be utilized for the clustering of retinal vessel in fundus images. (2) From real data: Corresponding annotations of target structure can be directly set as the prior knowledge. (3) Simulation: Based on the comprehension, experts can draw the pseudo masks to show the information of relative location, topology, etc. For example, based on retinal vessel mask, the approximate location of OD and fovea can be identified. Then, ellipses can be placed at these positions to represent OD and fovea based on their geometry priors.

\section{Experiments and Discussion}
The experiments can be divided into two parts to show the effectiveness of our proposed unsupervised local discrimination learning and prove the feasibility of combining local discrimination and topological priors to cluster target structures.

\subsection{Network architectures and initialization}
\label{Patch-wise}
The backbone is a U-net consisted with a VGG-liked encoder and a decoder. The encoder is a tiny version of VGG-16 without fully connection layers (FCs), whose channel number is quarter of VGG-16. The decoder is composed with 4 convolution blocks, each of which is made up of two convolution layers. The final features of decoder will be concatenated with the features generated by the first convolution block of encoder for the further processing of the clustering branch and the embedding branch. Embedding branch is formed of 2 convolution layers with 32 channels and a $l_2$ normalization layer to project each pixel onto a 32-D hypersphere. Clustering branch is consisted with 2 convolution layers with 8 channels followed by a $l_1$ normalization layer.

To minimize the KL divergence between the pseudo segmentation distribution and the references distribution, a discriminator is created. The discriminator is a simple classifier with 7 convolution layers and 2 FCs. The channel numbers of convolution layers are 16, 32, 32, 32, 32, 64, 64 and the first 5 layers are followed by a max-pooling layer to halve the image size. FCs' channels are 32 and 1, and the final FC is followed by a Sigmoid layer.

\textbf{Patch discrimination to initialize the network:} It is hard to simultaneously train the clustering branch and the embedding branch from scratch. Thus, we firstly jointly pre-train the backbone and the embedding branch by patch discrimination, which is an improvement of instance discrimination \cite{ye2019unsupervised}. The main idea is that the embedding branch should project similar patches (patches under various augmentations) onto close positions on the hypersphere. The embedding features $v_{i}$ will be firstly processed by an adaptive average pooling layer (APP) to generate spatial features, each of which represents feature of corresponding patches of image $x_{i}$. We denote $s_{i}(j)$ as the embedding vector for $x_{i}(j)$ ($j$-th patch of $x_{i}$), where $||s_{i}(j)||_{2}=1$ by applying a $l_2$ normalization. $\hat{s}_{i}(j)$ denotes the embedding vector of corresponding augmentation patch $\hat{x}_{i}(j)$. The probability of region $\hat{x}_{i}(j)$ being recognized as region ${x}_{i}(j)$ can be defined as followed:
\begin{equation}
	\begin{aligned}
		P(ij|\hat x_{i}(j))&=\frac{\exp(s_{i}^{T}(j)\hat s_{i}(j)/\tau)}{\sum_{k,l}{\exp(s_{k}^T(l)\hat s_{i}(j)/\tau)}},
	\end{aligned}
	\label{prob}
\end{equation}
Assuming all patches being recognized as $x_{i}(j)$ is independent, then the joint probability of $\hat x_{i}(j)$ being recognized as $x_{i}(j)$ and $x_{k}(l) (k\neq{i}\ or\ l\neq{j})$ not being recognized as $x_{i}(j)$ is as followed:
\begin{equation}
	P_{ij}=P(ij|\hat{x}_{i}(j))\prod_{k\neq i\ or\ l\neq j}(1-P(ij|\hat{x}_{k}(l))).
\end{equation}
The negative log likelihood and loss function are formulated as followed:
\begin{equation}
	J_{ij}=-\log P(ij|\hat{x}_{i}(j))-\sum_{k\neq i\ or\ l\neq j}\log(1-P(ij|\hat{x}_{k}(l))).
\end{equation}
\begin{equation}
	loss_{pd}=\sum_{i,j}J_{ij}
\end{equation}
We also introduce mixup \cite{zhang2018mixup} to make the representations more robust. Based on mixup, virtual sample $\tilde{x}_{i}=\lambda x_{a}+(1-\lambda)x_{b}$ is firstly generated by linear interpolation of $x_{a}$ and $x_{b}$, where $\lambda \in (0,1)$. The embedded representation for patch $\tilde{x}_{i}(j)$ is $\tilde{s}_{i}(j)$, and we expect it is similar to the mixup feature $z_{i}(j)$. The loss is defined as followed:
\begin{equation}
	\begin{aligned}
		z_{i}(j)=\frac{\lambda s_{a}(j)+(1-\lambda) s_{b}(j)}{||\lambda s_{a}(j)+(1-\lambda) s_{b}(j)||_2}.
	\end{aligned}
\end{equation}
\begin{equation}
	\tilde{P}(ij|\tilde{s}_{i}(j))=\frac{\exp(z_{i}^T(j)\tilde{s}_{i}(j)/\tau)}{\sum_{k,l}{\exp(z_{k}^T(l)\tilde{s}_{i}(j)/\tau)}}.
\end{equation}
\begin{equation}
	\tilde{J}_{ij}=-\log\tilde{P}(ij|\tilde{s}_{i}(j))-\sum_{k\neq i\ or\ l\neq j}\log(1-\tilde{P}(ij|\tilde{s}_{k}(l))),
\end{equation}
\begin{equation}
	loss_{mixup}=\sum_{i,j}\tilde{J}_{ij}
\end{equation}
When pre-training this model, we set the training loss as $loss_{pd}+loss_{mixup}$. The output size of APP is set as $4\times 4$ to split each image into 16 patches. And each batch contains 16 groups of images and 8 corresponding mixup images, and each of group contains 2 augmentations of one image. The augmentation methods contain $RandomResizedCrop$, $RandomGrayscale$, $ColorJitter$, $RandomHorizontalFlip$, $Rotation90$ in pytorch. The optimizer is Adam with initial learning rate ($lr$) of $0.001$, which will be half if the validation loss does not decrease over 3 epochs. The maximum training epoch is 20.
\subsection{Experiments for learning local discrimination}
\subsubsection{Datasets and preprocessing:} 
Our method is evaluated in two medical scenes. \textbf{Fundus images}: The model will be firstly trained on diabetic retinopathy (DR) detection dataset of kaggle \cite{cuadros2009eyepacs}\footnote{https://www.kaggle.com/c/diabetic-retinopathy-detection/data} ($30k$ for training, $5k$ for validation). Then, the pre-trained encoder is transferred to 8 segmentation tasks: (1) Retinal vessel: DRIVE \cite{staal2004ridge} (20 for training, 20 for testing), STARE \cite{hoover2000locating} (10 for training, 10 for testing) and CHASEDB1 \cite{owen2009measuring} (20 training, 8 testing). (2) OD and cup: Drishti-GS \cite{sivaswamy2014drishti} (50 for training, 50 for testing). ID (OD) \cite{porwal2018indian} (54 for training, 27 for testing). (3) Lesions: Haemorrhages dataset (Hae) and hard exudates dataset (HE) from IDRID \cite{porwal2018indian}. \textbf{Chest X-ray:} The encoder is pre-trained on ChestX-ray8 \cite{wang2017chestx} ($100k$ for training and $12k$ for validation) and transferred to lung segmentation \cite{candemir2013lung} (69 for training, 69 for testing). All images of above datasets are resized to $512\times512$.

\subsubsection{Implementation details:}
(1) Local discriminative representation learning: The model is firstly initialized by pre-trained model of patch discrimination. Then the training loss is set as $loss_{pd}+loss_{mixup}+10loss_{ld}+loss_{entropy}+5loss_{area}$. Each batch has 6 groups of images, each of which contains 2 augmentations of one image, and 3 mixup images. The maximum training epoch is 80 and the optimizer is Adam with $lr=0.001$.

(2) Transferring: The encoder of downstream tasks is initialized by the learnt feature extractor of local discrimination. The decoder is composed with 5 convolution blocks, each of which contains 2 convolution layers and is followed by a up-pooling layer. The loss is set as $loss_{dsc}=\frac{2|p\times g|}{|g|+|p|}$. This model will be firstly trained in 100 epochs in frozen pattern with Adam with $lr=0.001$, and then be trained in fine-tune pattern with $lr=0.0001$ in the following 100 epochs.

(3) Comparative methods: \textbf{Random:} The network is trained from scratch. \textbf{Supervised:} Supervised by the manual score of DR, the encoder will be firstly trained by making classification. \textbf{Wu et al. \cite{wu2018unsupervised}} and \textbf{Ye et al. \cite{ye2019unsupervised}:} Instance discrimination methods proposed in \cite{wu2018unsupervised} and \cite{ye2019unsupervised}. \textbf{LD:} The proposed method.

\begin{table}[t]
	\caption{Comparison of results of downstream segmentation tasks.}
	\centering
	\renewcommand\tabcolsep{2pt}
	\label{table:comparative experiments fundus}
	\begin{tabular}{l|ccc|ccc|cc|c}
		\toprule[0.5pt]
		&\multicolumn{3}{|c|}{Retinal vessel}&
		\multicolumn{3}{|c|}{Optic disc and cup}
		&\multicolumn{2}{|c}{Lesions }
		&\multicolumn{1}{|c}{X-ray }\\
		\hline
		encoder& DRIVE & STARE & CHASE & GS(cup) & GS(OD) & ID(OD)& Hae & HE & Lung\\				
		\hline
		\hline
		Random
		& $80.76$ & $76.26$ & $78.30$
		& $77.76$ & $95.41$ & $89.11$
		& $37.76$ & $57.44$ & $96.34$\\
		
		Supervised
		& $81.06$ & $80.59$ & $78.56$ 
		& $86.94$ & $96.40$ & $93.56$
		& $\textbf{51.11}$ & $61.34$ & $-$\\
		
		Wu et al.
		& $74.98$ & $66.15$ & $68.31$
		& $84.59$ & $94.58$ & $88.70$
		& $26.34$ & $48.67$ & $96.27$\\
		
		Ye et al.
		& $80.87$ & $81.22$ & $79.85$ 
		& $87.30$ & $\textbf{97.40}$ & $94.68$
		& $46.79$ & $59.40$ & $96.63$\\

		\hline
		LD
		& $\textbf{82.15}$ & $\textbf{83.42}$ & $\textbf{80.35}$ 
		& $\textbf{89.30}$ & $96.53$ & $\textbf{95.59}$
		& $46.72$ & $\textbf{65.77}$ & $\textbf{97.51}$ \\
		\hline
		
	\end{tabular}
\end{table}
		
\subsubsection{Results:}
The evaluation metric is mean Dice-Sørensen coefficient (DSC): $DSC=\frac{2{|P\times{G}|}}{|P|+|G|}$, where $P$ is the binary results of predictions and $G$ is the ground truth. Quantitative evaluations for downstream tasks are shown in Table \ref{table:comparative experiments fundus}, and we can have following observations:

1) The generalization ability of the trained local discriminative representation is demonstrated by the leading performance in the 6 fundus tasks and lung segmentation. Compared with models trained from scratch, models initialized by our pre-trained encoder can respectively gain improvements of 1.39\%, 7.16\%, 2.05\%, 11.54\%, 1.12\%, 6.48\%, 8.96\%, 8.33\% and 1.17\% in DSC for all 9 tasks.

2) Compared with instance discrimination methods by Wu et al. \cite{wu2018unsupervised} and Ye et al. \cite{ye2019unsupervised}, the proposed local discrimination is capable to learn finer features and is more suitable for unsupervised representation learning of medical images.

3) The proposed unsupervised method is free from labeled images and the learnt representation is more generalized, while supervised representation learning relies on expensive manual annotations and learns specialized representations. As shown in Table \ref{table:comparative experiments fundus}, our method shows better performance than supervised representation learning, whose target is to classification DR, and the only exception is on segmenting haemorrhages which is the key evidence for DR. 

\begin{figure*}[t]
	\centering	
	\includegraphics[width=\textwidth]{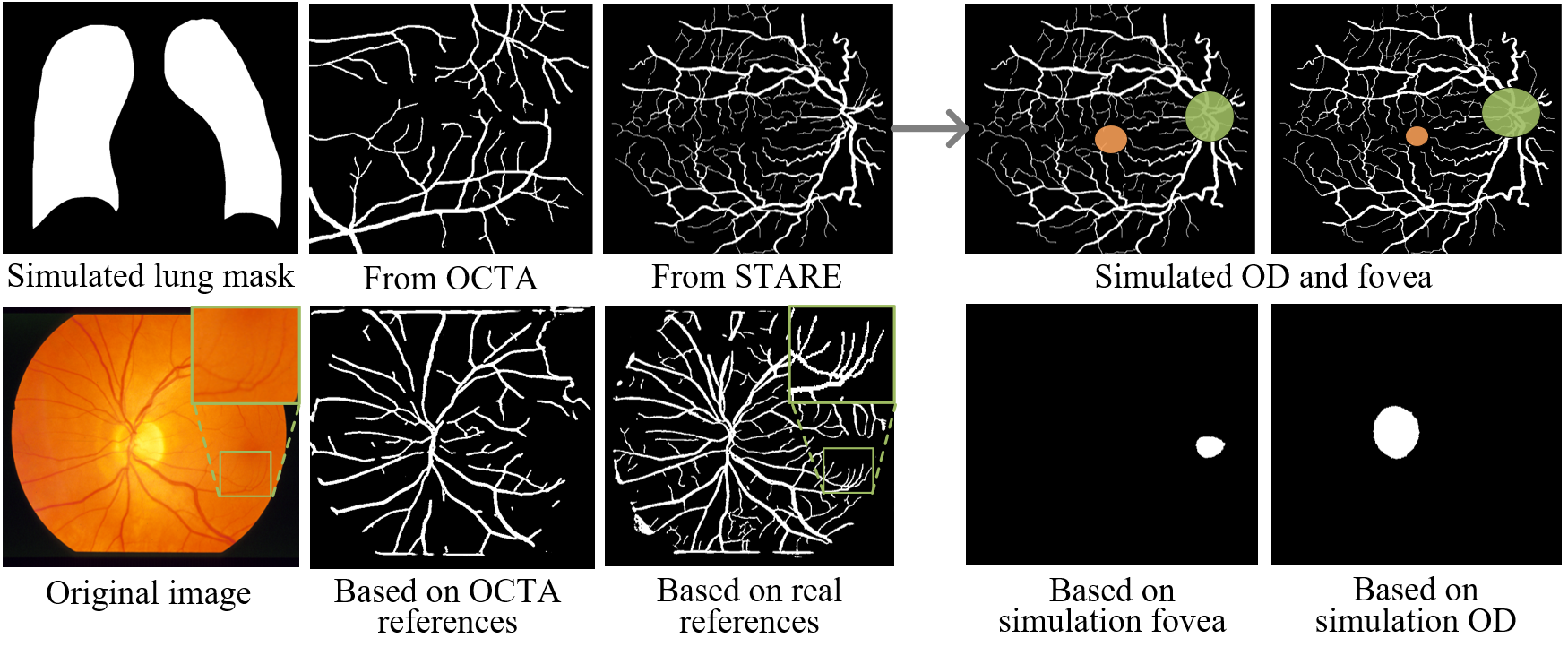}
	\caption{Some examples of reference masks and predicted results: The first row show some reference images and the second row show the predictions.} 
	\label{fig:prior_guide}
\end{figure*}
\subsection{Experiments for clustering structures based on prior knowledge}
\subsubsection{Implementation details:}
In this part, we respectively fuse reference images from real data, similar structures and simulations into local discrimination to investigate the ability of clustering anatomical structures. A dataset with 3110 high-quality fundus images from \cite{wang2018weakly-supervised} and 1482 frontal X-rays from \cite{wang2017chestx} are utilized as the training data. The reference images can be constructed in 3 ways: (1) From real references: ALL 40 retinal vessel masks of DRIVE are utilized as the references for clustering pixels of vessel. (2) From similar structures: Similar structures share similar priors, thus, 10 OCTA vessel masks are utilized as the references for retinal vessel of fundus. (3) Simulation: We directly draw 20 simulated lung masks to guide lung segmentation. Meanwhile, based on vessel masks of DRIVE, we place ellipses at approximate center location of OD and fovea to generate pseudo masks. Some reference masks are shown in Figure \ref{fig:prior_guide}.

$f_{\theta}$ needs to jointly learn local discrimination and cheat $D$, thus, it will be updated by minimizing the following loss:
\begin{equation}
	loss_{f_{\theta}}=loss_{pd}+loss_{mixup}+10loss_{ld}+loss_{entropy}+5loss_{area}+2loss_{adv}
\end{equation}
The optimizer for $f_{\theta}$ is Adam with $lr=0.001$. The discriminator is optimized by minimizing $loss_D$ and the optimizer is Adam with $lr=0.0005$. It is worth noting that during the clustering training of OD and fovea, all masks of real vessel, fovea and OD are concatenated and fed into $D$ to provide enough information and $f_{\theta}$ is firstly pre-trained to cluster retinal vessel. The maximum training epoch is 80.

\subsubsection{Results:} Visualization examples are shown in Figure \ref{fig:prior_guide}. Quantitative evaluations are as followed: (1) Retinal vessel segmentation is evaluated in the test data of STARE. And the $DSC$ are respectively 66.25\% and 57.35\% for models based on real references and based on OCTA annotations. (2) The segmentation of OD is evaluated in the test data of Drishti-GS and gains $DSC$ of 83.60\%. (3) The segmentation of fovea is evaluated in the test data of STARE. Because the region of fovea is fuzzy, we measure the mean distance between the real center of fovea and the predicted center. The mean distance is $7.63 pixels$. (4) The segmentation of lung is evaluated in NLM \cite{candemir2013lung} and the DSC is 81.20\%. 

Based on above results, we can have following observations: 

1) In general, topological priors generated from simulation or similar structures in a different modality is effective to guide the clustering of target regions.

2) However, real masks contain more detailed information and are able to provide more precise guidance. For example, compared with vessel segmentations based on OCTA annotations, which missing the thin blood vessels due to the great thickness of OCTA mask, segmentations based on real masks can recognize thin vessels due to the details provided and the constraint of clustering pixels with similar context. 

3) For anatomical structures with fuzzy intensity pattern, such as fovea, combining local similarity and structure priors is able to guide precise recognition.

\section{Conclusion}
In this paper, we propose an unsupervised framework to learn local discriminative representation for medical images. By transferring the learnt feature extractor, downstream tasks can be improved to decrease the demand for expensive annotations. Furthermore, similar structures can be clustered by fusing prior knowledge into the learning framework. The experimental results show that our methods have best performance on 7 out of 9 tasks in fundus and chest X-ray images, demonstrating the great generalization of the learnt representation. Meanwhile, the feasibility of clustering structures based on prior knowledge and unlabeled images is demonstrated by combining local discrimination and topological priors from real data, similar structures or even simulations to segment anatomical structures including retinal vessel, OD, fovea and lung.

\bibliographystyle{splncs04}
\bibliography{mybibfile}

\begin{thebibliography}{10}
\providecommand{\url}[1]{\texttt{#1}}
\providecommand{\urlprefix}{URL }
\providecommand{\doi}[1]{https://doi.org/#1}

\bibitem{bachman2019learning}
Bachman, P., Hjelm, R.D., Buchwalter, W.: Learning representations by
  maximizing mutual information across views. In: Advances in Neural
  Information Processing Systems. pp. 15535--15545 (2019)

\bibitem{candemir2013lung}
Candemir, S., Jaeger, S., Palaniappan, K., Musco, J.P., Singh, R.K., Xue, Z.,
  Karargyris, A., Antani, S., Thoma, G., McDonald, C.J.: Lung segmentation in
  chest radiographs using anatomical atlases with nonrigid registration. IEEE
  transactions on medical imaging  \textbf{33}(2),  577--590 (2013)

\bibitem{cuadros2009eyepacs}
Cuadros, J., Bresnick, G.: Eyepacs: an adaptable telemedicine system for
  diabetic retinopathy screening. Journal of diabetes science and technology
  \textbf{3}(3),  509--516 (2009)

\bibitem{dosovitskiy2015discriminative}
Dosovitskiy, A., Fischer, P., Springenberg, J.T., Riedmiller, M., Brox, T.:
  Discriminative unsupervised feature learning with exemplar convolutional
  neural networks. IEEE transactions on pattern analysis and machine
  intelligence  \textbf{38}(9),  1734--1747 (2015)

\bibitem{glorot2011deep}
Glorot, X., Bordes, A., Bengio, Y.: Deep sparse rectifier neural networks. In:
  Proceedings of the fourteenth international conference on artificial
  intelligence and statistics. pp. 315--323 (2011)

\bibitem{goodfellow2014generative}
Goodfellow, I., Pouget-Abadie, J., Mirza, M., Xu, B., Warde-Farley, D., Ozair,
  S., Courville, A., Bengio, Y.: Generative adversarial nets. In: Advances in
  neural information processing systems. pp. 2672--2680 (2014)

\bibitem{he2020momentum}
He, K., Fan, H., Wu, Y., Xie, S., Girshick, R.: Momentum contrast for
  unsupervised visual representation learning. In: Proceedings of the IEEE/CVF
  Conference on Computer Vision and Pattern Recognition. pp. 9729--9738 (2020)

\bibitem{hoover2000locating}
Hoover, A., Kouznetsova, V., Goldbaum, M.: Locating blood vessels in retinal
  images by piecewise threshold probing of a matched filter response. IEEE
  Transactions on Medical imaging  \textbf{19}(3),  203--210 (2000)

\bibitem{mahmood2020whole}
Mahmood, U., Rahman, M.M., Fedorov, A., Lewis, N., Fu, Z., Calhoun, V.D., Plis,
  S.M.: Whole milc: generalizing learned dynamics across tasks, datasets, and
  populations. In: International Conference on Medical Image Computing and
  Computer-Assisted Intervention. pp. 407--417. Springer (2020)

\bibitem{owen2009measuring}
Owen, C.G., Rudnicka, A.R., Mullen, R., Barman, S.A., Monekosso, D., Whincup,
  P.H., Ng, J., Paterson, C.: Measuring retinal vessel tortuosity in
  10-year-old children: validation of the computer-assisted image analysis of
  the retina (caiar) program. Investigative ophthalmology \& visual science
  \textbf{50}(5),  2004--2010 (2009)

\bibitem{porwal2018indian}
Porwal, P., Pachade, S., Kamble, R., Kokare, M., Deshmukh, G., Sahasrabuddhe,
  V., Meriaudeau, F.: Indian diabetic retinopathy image dataset (idrid): a
  database for diabetic retinopathy screening research. Data  \textbf{3}(3),
  ~25 (2018)

\bibitem{sivaswamy2014drishti}
Sivaswamy, J., Krishnadas, S., Joshi, G.D., Jain, M., Tabish, A.U.S.:
  Drishti-gs: Retinal image dataset for optic nerve head (onh) segmentation.
  In: 2014 IEEE 11th international symposium on biomedical imaging (ISBI). pp.
  53--56. IEEE (2014)

\bibitem{staal2004ridge}
Staal, J., Abr{\`a}moff, M.D., Niemeijer, M., Viergever, M.A., Van~Ginneken,
  B.: Ridge-based vessel segmentation in color images of the retina. IEEE
  transactions on medical imaging  \textbf{23}(4),  501--509 (2004)

\bibitem{wang2018weakly-supervised}
Wang, R., Chen, B., Meng, D., Wang, L.: Weakly-supervised lesion detection from
  fundus images. IEEE transactions on medical imaging pp. 1501--1512 (2018)

\bibitem{wang2017chestx}
Wang, X., Peng, Y., Lu, L., Lu, Z., Bagheri, M., Summers, R.M.: Chestx-ray8:
  Hospital-scale chest x-ray database and benchmarks on weakly-supervised
  classification and localization of common thorax diseases. In: Proceedings of
  the IEEE conference on computer vision and pattern recognition. pp.
  2097--2106 (2017)

\bibitem{wu2018unsupervised}
Wu, Z., Xiong, Y., Yu, X.S., Lin, D.: Unsupervised feature learning via
  non-parametric instance discrimination. Proceedings of the IEEE Conference on
  Computer Vision and Pattern Recognition pp. 3733--3742 (2018)

\bibitem{xie2020instance}
Xie, X., Chen, J., Li, Y., Shen, L., Ma, K., Zheng, Y.: Instance-aware
  self-supervised learning for nuclei segmentation. In: International
  Conference on Medical Image Computing and Computer-Assisted Intervention. pp.
  341--350. Springer (2020)

\bibitem{ye2019unsupervised}
Ye, M., Zhang, X., Yuen, P.C., Chang, S.F.: Unsupervised embedding learning via
  invariant and spreading instance feature. In: Proceedings of the IEEE
  Conference on computer vision and pattern recognition. pp. 6210--6219 (2019)

\bibitem{zhang2018mixup}
Zhang, H., Cisse, M., Dauphin, Y.N., Lopez-Paz, D.: mixup: Beyond empirical
  risk minimization. In: International Conference on Learning Representations
  (2018)

\end{thebibliography}
\end{document}